\documentclass[runningheads]{llncs}

\usepackage{eccv}

\usepackage{eccvabbrv}

\usepackage{graphicx}
\usepackage{booktabs}
\usepackage{adjustbox}

\usepackage[accsupp]{axessibility}  % Improves PDF readability for those with disabilities.

\usepackage{hyperref}

\usepackage{orcidlink}

\begin{document}

\title{Light Field Diffusion for Single-View Novel View Synthesis} 


\author{Yifeng Xiong\inst{1} \and
Haoyu Ma\inst{1} \and
Shanlin Sun\inst{1} \and
Kun Han\inst{1} \and
Hao Tang\inst{1} \and
Xiaohui Xie\inst{1}}

\authorrunning{Xiong et al.}

\institute{University of California, Irvine \\
\email{\{yifengx4, haoyum3, shanlins, khan7, htang6, xhx}@uci.edu\}}

\maketitle
\begin{center}
  \includegraphics[width=\linewidth]{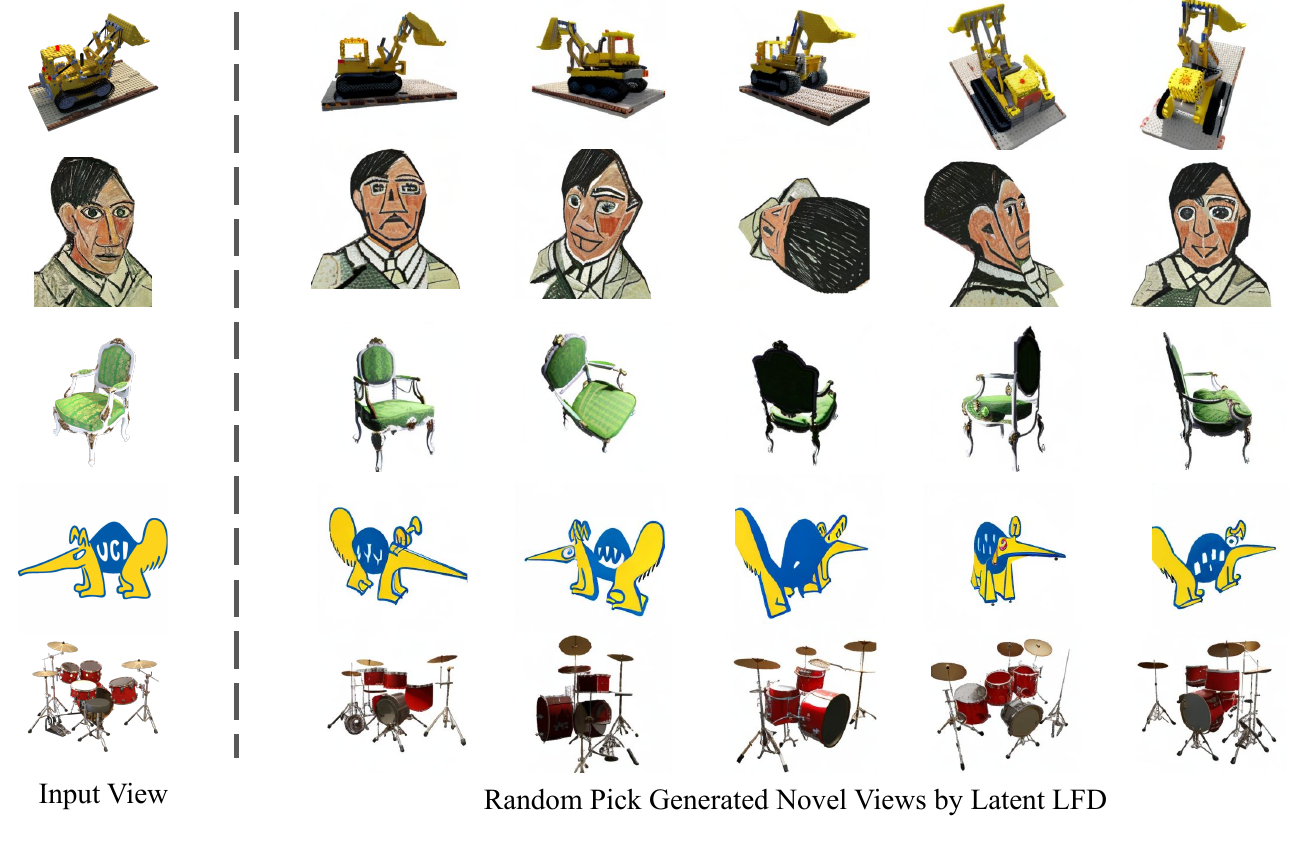}
  \captionof{figure}{\textbf{Zero-shot single-view novel view synthesis by Latent Light Field Diffusion.} Given a single input view, our method can generate novel views from various viewpoints while maintaining consistency with the reference image.}
  \label{fig:teaser}
\end{center}

\begin{abstract}
    Single-view novel view synthesis (NVS), the task of generating images from new viewpoints based on a single reference image, is important but challenging in computer vision. Recent advancements in NVS have leveraged Denoising Diffusion Probabilistic Models (DDPMs) for their exceptional ability to produce high-fidelity images. However, current diffusion-based methods typically utilize camera pose matrices to globally and implicitly enforce 3D constraints, which can lead to inconsistencies in images generated from varying viewpoints, particularly in regions with complex textures and structures.

    To address these limitations, we present Light Field Diffusion (LFD), a novel conditional diffusion-based approach that transcends the conventional reliance on camera pose matrices. Starting from the camera pose matrices, LFD transforms them into light field encoding, with the same shape as the reference image, to describe the direction of each ray. By integrating light field encoding with the reference image, our method imposes local pixel-wise constraints within the diffusion process, fostering enhanced view consistency. Our approach not only involves training image LFD on the ShapeNet Car dataset but also includes fine-tuning a pre-trained latent diffusion model on the Objaverse dataset. This enables our latent LFD model to exhibit remarkable zero-shot generalization capabilities across out-of-distribution datasets like RTMV as well as in-the-wild images. Experiments demonstrate that LFD not only produces high-fidelity images but also achieves superior 3D consistency in complex regions, outperforming existing novel view synthesis methods. Project page can be found at \url{https://lightfielddiffusion.github.io/}.
\end{abstract}

\section{Introduction}
\label{sec:Introduction}

Novel view synthesis (NVS), the inference of a 3D scene's appearance from novel viewpoints given several images of the scene \cite{watson2022novel, park2017transformation, peng2021neural, tretschk2021non, flynn2019deepview, zhou2018stereo}, plays a fundamental role in many computer vision applications such as game studios, virtual reality, and augmented reality \cite{musialski2013survey, chen1995quicktime, szeliski2023creating}. 
This task requires that synthesized images not only match specific novel viewpoints but also ensure shapes and details are consistent with reference images, i.e., view consistency.

% NeRF 
Recently, Neural Radiance Field (NeRF) \cite{mildenhall2021nerf} has made significant advancements in the area of novel view synthesis. 
NeRF represents a scene using a multi-layer perception (MLP) and uses volume rendering to generate an image, enabling realistic 3D renderings and multi-view consistency. 
In spite of its great success, the training process of NeRF typically requires dense views from different angles and corresponding camera poses, posing a challenge for its practical application in real-world scenarios. To this end, later works focus on learning a NeRF in the sparse view settings \cite{yu2021pixelnerf,niemeyer2022regnerf}, which either utilize prior knowledge from pre-trained models on large-scale datasets \cite{yu2021pixelnerf,lin2023vision} or introduce regularization on geometry \cite{niemeyer2022regnerf}. 
Moreover, some recent works \cite{xu2022sinnerf,lin2023vision} try to train NeRFs to generate new images from different views with only one reference image, i.e., single-view novel view synthesis. This task is exceptionally demanding for regression-based methods like NeRF \cite{watson2022novel}. Without sufficient information from other views, NeRF suffers from incorrect learning in 3D geometry, which often results in blurred outputs \cite{niemeyer2022regnerf}.

\begin{figure}
  \centering
   \includegraphics[width=1.\linewidth]{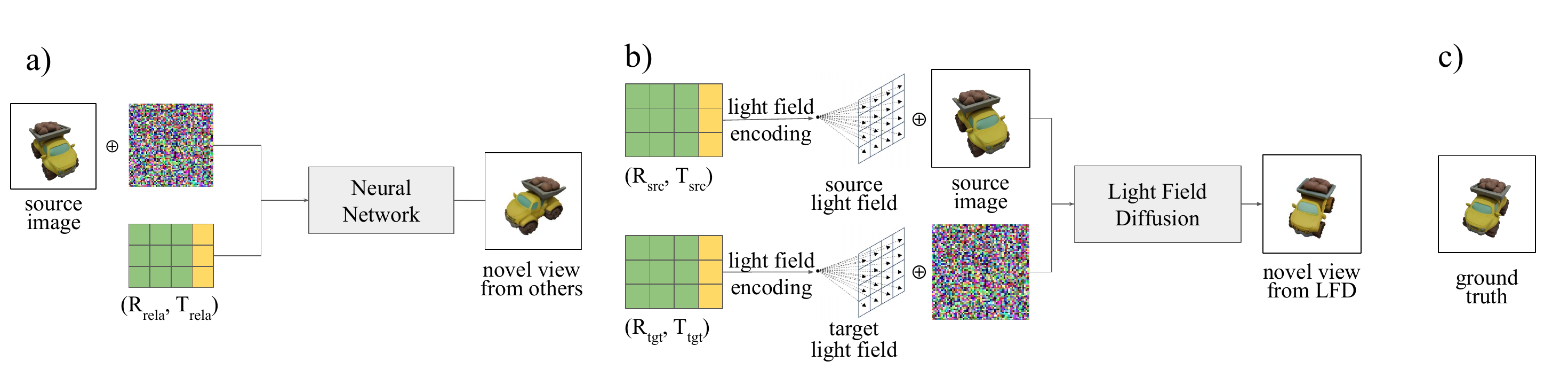}
   \caption{\textbf{Comparison of our Light Field Diffusion and previous diffusion-based models for single-view novel view synthesis:}
   a) Previous models \cite{watson2022novel,liu2023zero1to3} directly take camera pose matrices (rotation R and translation T) as input, which can only provide 3D constraints globally and implicitly. 
   b) Our Light Field Diffusion transforms the camera pose matrices into light field encoding and concatenates them with the noise and source image, which provides local and explicit pixel-wise 3D geometry constraints, enabling better novel view consistency. 
   }
   \label{fig:overall_compare}
\end{figure}

On the other hand, generative methods, especially Diffusion Probabilistic Model (DDPM) \cite{ho2020denoising, song2019generative, sohl2015deep}, have become popular in the field of novel view synthesis \cite{liu2023zero1to3, chan2022efficient, xu2023neurallift,watson2022novel, zhu2018generative, rombach2021geometry} due to their strong ability in single-view novel view synthesis with large view rotations.  
Of these approaches, one line of research combines the merits of NeRF with diffusion models \cite{xu2023neurallift,gu2023nerfdiff} and shows that these two types of methods can guide and boost each other. 
Other approaches directly train a conditional diffusion model in an end-to-end manner. Notably, 3DiM \cite{watson2022novel} is a pioneering work that learns a diffusion model conditioned on camera pose matrices of both views. 
In subsequent developments, Zero-1-to-3 \cite{liu2023zero1to3} accomplishes this by finetuning the pre-trained latent diffusion model \cite{rombach2022high} using relative camera rotation and translation as conditions. This approach leverages the ample 3D priors inherent in large-scale diffusion models, despite being trained primarily on 2D images.

Although these end-to-end methods achieve great performance, they usually ignore the fact that the camera pose matrices can only provide 3D constraints globally and implicitly. 
As a result, these methods usually require a heavy network with billions of parameters to learn the implicit 3D constraint. 
Moreover, the generated images in different views may suffer pixel-wise inconsistency with the reference image. 
Therefore, it is imperative to establish an effective camera pose representation within these end-to-end diffusion-based methods. 

To address the above issue, we propose Light Field Diffusion (LFD), a novel framework for single-view novel view synthesis based on diffusion models \cite{ho2020denoising, rombach2022high}.
Instead of relying directly on camera pose matrices, LFD takes a different approach by converting these matrices into pixel-wise camera pose encodings based on the light field \cite{levo1996light,sitzmann2021lfns}, 
which includes the ray direction of each pixel and the camera origin. 
In detail, we concatenate this light field encoding with both target image and reference image to involve camera pose information and then use cross-attention layers to model the relationship between the reference and target. This design facilitates the establishment of local pixel correspondences between the reference and target images, ultimately enhancing novel view consistency in the generated images. \cref{fig:overall_compare} summarizes an overall comparison between our method and previous end-to-end diffusion-based models in novel view synthesis. 

We summarize our contributions as follows:
\begin{itemize}

    \item We present a novel framework called Light Field Diffusion, an end-to-end conditional diffusion model designed for synthesizing novel views using a single reference image. Instead of directly using camera pose matrices, we employ a transformation into pixel-wise light field encoding. This approach harnesses the advantages of local pixel-wise constraints, resulting in a significant enhancement in model performance.

    \item We train the Light Field Diffusion on both latent space and image space. In latent space, we finetune a pre-trained latent diffusion model on the Objaverse dataset \cite{Deitke_2023_CVPR}. In image space, we train a conditional DDPM on the ShapeNet Car dataset\cite{chang2015shapenet}.

    \item Our method is demonstrated to effectively synthesize novel views that maintain consistency with the reference image, ensuring viewpoint coherence and producing high-quality results. Our latent Light Field Diffusion model also showcases the exceptional ability for zero-shot generalization to out-of-distribution datasets such as RTMV \cite{tremblay2022rtmv} and in-the-wild images.

\end{itemize}

\section{Related Work}
\label{sec:Related_work}

\subsection{NeRF in Novel View Synthesis}
Neural Radiance Fields (NeRFs) \cite{mildenhall2021nerf, yu2021pixelnerf, niemeyer2022regnerf, lin2023vision, xu2022sinnerf, jain2021putting, jang2021codenerf, schwarz2020graf} have exhibited promising results in the realm of novel view synthesis. In detail, NeRFs employ a fully connected neural network to establish a mapping from viewing direction and spatial location to RGB color and volume density, enabling the rendering of views by projecting rays into 3D space and querying the radiance field at various points along each ray for each pixel on the image plane. Consequently, NeRF can guarantee 3D consistency among generated images. 
However, the vanilla NeRF is a scene-specific model, and it usually requires a dense view sampling of the scene, which may be impractical for complicated scenes.
Since then, there have been multiple following works that improve the quality and generalization ability of NeRFs \cite{yu2021pixelnerf,jain2021putting,niemeyer2022regnerf,lin2023vision,chen2021mvsnerf,trevithick2021grf,xu2022sinnerf}. 

For instance, 
PixelNeRF \cite{yu2021pixelnerf} uses a CNN feature extractor pre-trained on large-scale datasets to extract context information features and conduct NeRF in feature map space. 
VisionNeRF \cite{lin2023vision} integrates both global features from a pre-trained ViT \cite{dosovitskiy2010image} model and local features from pre-trained CNN encoders. 
DietNeRF \cite{jain2021putting} supervises the training process by a semantic consistency loss using the CLIP encoder \cite{radford2021learning}.
Most recently, researchers have pushed NeRF to single-view novel view synthesis \cite{yu2021pixelnerf, xu2022sinnerf}. 
SinNeRF \cite{xu2022sinnerf} resorts to the depth map of the single reference image and proposes a semi-supervised framework to guide the training process of NeRF. However, due to the properties of regressive models, all these NeRF-based approaches suffer from blurred results when the view rotation is large \cite{niemeyer2022regnerf}.

\subsection{Light Field Rendering}
Light field \cite{levo1996light} is an alternative representation of 3D scene and has achieved competitive results in few-view novel view synthesis \cite{sitzmann2021lfns, sajjadi2022scene,suhail2022light,suhail2022generalizable}. 
Instead of encoding a scene in 3D space, light field maps an oriented camera ray to the radiance observed by that ray. During the rendering process, the network is only evaluated once per pixel \cite{sitzmann2021lfns}, rather than hundreds of evaluations in volume rendering. 
Thus, light field not only incorporates 3D constraints but also attains remarkable computational efficiency. 

Specifically, Light Field Network (LFN) \cite{sitzmann2021lfns} captures the combined essence of the geometry and appearance of the underlying 3D scene within a neural implicit representation, parameterized as a 360-degree, four-dimensional light field.
Scene Representation Transformers (SRT) \cite{sajjadi2022scene} is a transformer-based framework that considers light field rays as positional encoding.  It employs transformer encoders to encode source images and corresponding light field rays into set-latent representation and render novel view images by attending to the latent representation with light field rays in target views. 
Despite its efficiency, light field rendering also suffers from blurred outputs in single-view novel view synthesis due to regressive characteristics like NeRF. 
On the contrary, our method integrates the light field representation of the camera view with generative diffusion models, thus leading to high sample quality.

\subsection{Generative Models in Novel View Synthesis}

\subsubsection{GAN in NVS.}
Generative Adversarial Networks (GAN) \cite{goodfellow2014generative} play an important role in novel view synthesis. 
Early works consider novel view synthesis as an image-to-image framework \cite{sun2018multi} and apply adversarial training techniques to improve image quality.  
Recently, researchers have focused on building 3D-aware generative models \cite{gu2021stylenerf, niemeyer2021giraffe, chan2022efficient}. 
In detail, 
StyleNeRF \cite{gu2021stylenerf} integrates NeRF into a style-based generator \cite{karras2019style} to enable high-resolution image generation with multi-view consistency. 
EG3D \cite{chan2022efficient} introduces a hybrid explicit–implicit tri-plane representation to train a 3D GAN with neural rendering efficiently. Although these methods can generate high-fidelity images, they usually only work well in limited rotation angles. 

\subsubsection{Diffusion in NVS.}
Diffusion models \cite{ho2020denoising, song2019generative, sohl2015deep} have achieved state-of-the-art performance in computer vision community, including image generation \cite{rombach2022high, song2021scorebased, saharia2022palette, saharia2022photorealistic, nichol2021glide}, and likelihood estimation \cite{kingma2021variational, song2021maximum}. 
Recently, there are some works that apply diffusion to NVS \cite{zhou2023sparsefusion, anciukevivcius2023renderdiffusion, wynn2023diffusionerf, deng2023nerdi, gu2023nerfdiff, xu2023neurallift,tseng2023consistent, muller2023diffrf, chan2023generative, szymanowicz2023viewset_diffusion, yu2023long, shi2023mvdream, karnewar2023holodiffusion}.
In the field of novel view synthesis, some works take diffusion models as guidance in the training of NeRF, while others aim to train end-to-end diffusion models condition on reference images and target camera views.

Specifically, 3DiM \cite{watson2022novel} is a diffusion-based approach that is trained on pairs of images with corresponding camera poses. They also introduce the stochastic conditioning sampling algorithm, depending on the previous output to sample autoregressively, which can then sample an entire set of 3D-consistent outputs.
Zero-1-to-3 \cite{liu2023zero1to3} finetunes the large pre-trained latent diffusion model, Stable Diffusion \cite{rombach2022high}, with the source image channel concatenated with the input to U-Net. The CLIP \cite{radford2021learning} embedding of the source image is also combined with relative camera rotation and translation, which is treated as a global condition during the denoising steps.
The pose-guided diffusion \cite{tseng2023consistent} proposes to introduce the source image and relative camera poses into diffusion models with epipolar-based cross-attention, as it is challenging to learn the correspondence between source and target views via concatenated inputs. 
RenderDiffusion \cite{anciukevivcius2023renderdiffusion} trains the diffusion models with tri-plane representations from posed 2D images to enable 3D understanding, which can perform novel view synthesis and 3D reconstruction. 
However, all of these end-to-end approaches are conditioned on camera pose matrices, which may lead to the lack of pixel-wise multi-view consistency because camera pose matrices only provide  3D constraints implicitly and globally.

\section{Methodology}
\label{sec:Methdology}

\begin{figure*}[t]
  \centering
    \includegraphics[width=\linewidth]{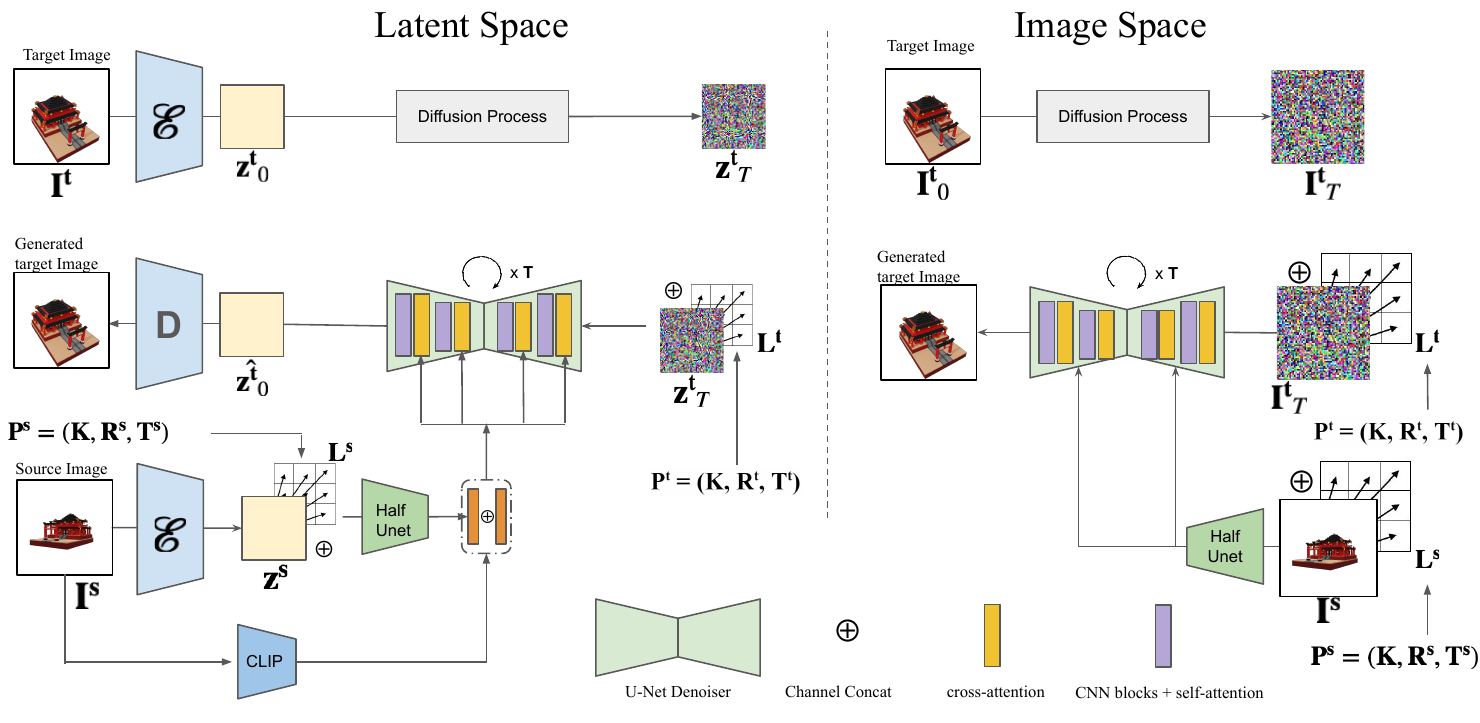}
    \caption{The overall pipeline of Ligh Field Diffusion (LFD) in both latent space and image space. The LFD translates the input source camera pose $\mathbf{P^s}$ and target camera pose $\mathbf{P^t}$ into source light field $\mathbf{L^s}$ and target light field $\mathbf{L^t}$. The U-Net denoiser takes the concatenation of noised image and $\mathbf{L^t}$ as inputs. The half U-Net extracts features from the concatenation of the source image and source light field $\mathbf{L^s}$ and interacts with the U-Net denoiser via cross-attention. }
  \label{fig:architecture}
\end{figure*}

\subsection{Overview}
In \cref{fig:architecture}, we provide an overview of our Light Field Diffusion (LFD) for novel view synthesis from a single reference. 
Given a single image of an object $\mathbf{I}^{s} \in \mathbb{R}^{H \times W \times 3}$ (source view) and its corresponding camera pose $\mathbf{P}^{s} = ( \mathbf{K}, \mathbf{R}^s, \mathbf{T}^s)$, the goal of our model $f(\cdot)$ is to generate a new image $\mathbf{I}^t \in \mathbb{R}^{H \times W \times 3}$ of the same object from a new viewpoint (target view) with camera pose $\mathbf{P}^t = (\mathbf{K}, \mathbf{R}^t, \mathbf{T}^t)$, where $H$ is image height, $W$ is image width,  $\mathbf{K} \in \mathbb{R}^{3\times3}$ is the intrinsic camera matrix, $\mathbf{R}^s$ and $\mathbf{R}^t \in \mathbb{R}^{3\times3}$ are rotation matrices, and $\mathbf{T}^s$ and $\mathbf{T}^t$ are translation matrices. 
In order to provide better pixel-wise constraints and view consistency, we first translate camera pose matrices into light field encoding $\mathbf{L} \in \mathbb{R}^{H\times W \times 180}$, which is a 2D representation of the camera pose matrices (\cref{subsec:light_field_encoding}). 
To enable the fidelity of generated images, we train a conditional Denoising Diffusion Probabilistic Model (DDPM) \cite{ho2020denoising} to synthesize the target image $\mathbf{I}^t $ with light field encodings for both source and target views. Thus, our LFD can be formulated as $\mathbf{I}^t = f(\mathbf{I}^{s}, \mathbf{L}^{s}, \mathbf{L}^{t})$ (\cref{subsec:lfd}). 

To demonstrate the benefits and generalization of LFD, we propose its implementations on both image pixel space and latent space (\cref{subsec:lfd_implement}). 
In detail, we first propose a framework based on the original DDPM \cite{ho2020denoising}, named Image LFD, which directly trains the light field diffusion in the image pixel space. 
Moreover, we propose a framework built upon the Latent Diffusion Model (LDM) \cite{rombach2022high}, named Latent Light Field Diffusion (Latent LFD), which can generate high-resolution images for any objects from novel views.

\subsection{Light Field Encoding}
\label{subsec:light_field_encoding}
In this section, we introduce how to translate the camera pose $\mathbf{P}$ into the 2D light field encoding $\mathbf{L}$ \cite{sitzmann2021lfns}, which is a 2D representation of the camera pose. 
Denote the coordinate of one pixel on the image plane as $(u,v)$, where $u \in \{1,2,...,h\}$ and $v \in \{1,2,...,w\}$. Here $(h,w)=(H/s, W/s)$, and $s$ is the downsample factor. We define a view ray $ \mathbf{r}_{uv} = (\mathbf{o}, \mathbf{d}_{uv}) \in \mathbb{R}^6$, where $\mathbf{o}$ is the coordinate of the camera center and $\mathbf{d}_{uv}$ is the direction vector. Under the world coordinate, the camera center $\mathbf{o}$ is equal to the translation vector  $\mathbf{T}$, and $\mathbf{d}_{uv}$ is defined by
\begin{equation}
    \mathbf{d}_{uv}= \mathbf{R}  \cdot  \mathbf{K}^{-1} \cdot [u, v, 1]^\intercal \in \mathbb{R}^3.
\end{equation}
Thus, the light field is defined by $L = ( {\mathbf{r}_{uv} ) }_{ h \times w } \in \mathbb{R}^{h \times w \times 6}$, which concatenates the origin field $^{(o)}L = ( \mathbf{o} )_{h\times w} \in \mathbb{R}^{h\times w \times 3}$ and the ray direction field $^{(d)}L = ( \mathbf{d}_{uv} )_{h \times w} \in \mathbb{R}^{h\times w \times 3}$. 
To simplify the learning process, we take the source view image as the canonical image, hence the light field of both source view and target view are expressed in the coordinate of $[ \mathbf{R}^s | \mathbf{T}^s ]$. 
Thus, the origin field of the source view $^{(o)}L^s$ is a zero matrix. 
Following \cite{mildenhall2021nerf}, we further apply the positional encoding function $\gamma$ to map the 6D light fields into a higher dimensional space, named light field encoding $\mathbf{L}$,  which is defined by:
\begin{equation}
    \mathbf{L} = \gamma(L) = [ \sin(2^0\pi L), \cos (2^0\pi L), ..., \sin(2^{K-1}\pi L), \cos(2^{K-1}\pi L)]
\end{equation}
, where $K$ is the number of octaves. We set it to $15$. Thus, the light field encoding $\mathbf{L}$ has a channel of $180$. We apply it to both source view and target view light field and obtain $\mathbf{L}^s$,$\mathbf{L}^t \in \mathbb{R}^{h \times w \times 180} $. 

In the task of novel view synthesis, a key consideration is the consistency with the input source view. With the help of light field encoding, for pixels that are near each other, there is a tendency for their ray directions to be closely aligned, resulting in similar representations.

\subsection{Light Field Diffusion Models}
\label{subsec:lfd}
Our light field diffusion is a kind of conditional diffusion model, which adds additional conditions within  DDPM \cite{ho2020denoising}. 
In detail, given one object, we randomly select two images $\mathbf{I^s}$ and $\mathbf{I^t}$ from different views. 
We define the origin image from the target view as $\mathbf{I^t}_0$. 
In the diffusion steps, we gradually add random Gaussian noise $\epsilon$ to $\mathbf{I^t}_0$ until it becomes pure noise. 
The DDPM \cite{ho2020denoising} proposes that we can directly get noised image $\mathbf{I^t}_t$ at any timestep $t $ from $\mathbf{I^t}_0$ with the reparameterization trick: 
\begin{equation}
q(\mathbf{I^t}_{t}|\mathbf{I^t}_{0})=\mathcal{N}(\mathbf{I^t}_t; \sqrt{\bar{\alpha_t}} \mathbf{I^t}_{0},(1-\bar{\alpha_t})I ), t \in \{1,2,...,T\}
\end{equation}
, where $T$ is the total number of time steps. 
The goal of our light field diffusion $\epsilon_\theta$ is to reconstruct the origin image $\mathbf{I^t}_0$ from $\mathbf{I^t}_T$ with $\mathbf{I^s}$, $\mathbf{L^s}$, and $\mathbf{L^t}$ as conditions. We denote all of these conditions as $c$. 
In detail, the training loss of our light field diffusion is defined as: 

\begin{equation}
L(\theta)=\mathbb{E}_{\mathbf{I^t}, c, \epsilon \sim \mathcal{N}(0,I) , t}[||\epsilon - \epsilon_\theta(\mathbf{I^t}_{t}, t, \mathbf{I^s}, \mathbf{L^s}, \mathbf{L^t})||^2]
\end{equation}

\subsection{Implementation}
\label{subsec:lfd_implement}

\subsubsection{Image Light Field Diffusion.}

We employ the U-Net denoiser \cite{ronneberger2015u, ho2020denoising} with self-attention modules \cite{vaswani2017attention} $\mathcal{F}(\cdot)$ in DDPM to build our light field diffusion in image pixel space, which directly take the RGB image as inputs. We name it Image LFD for simplicity.  
As shown in \cref{fig:architecture}, we concatenate the target light field encodings $\mathbf{L^t}$ with downsample rate $s=1$ and the noise target image $\mathbf{I^t}_t$ together and obtain multi-scale features $\{\mathbf{f}_i^t \} = \mathcal{F}([\mathbf{L^t}, \mathbf{I^t}_t])$ with the U-Net. 
With this design, the ray information can be encoded with each pixel on the noise target image. 

Meanwhile, we capture information from the source view image $\mathbf{I^s}$. The pioneer diffusion-based novel view synthesis networks \cite{watson2022novel,liu2023zero1to3} simply concatenate the source view image $\mathbf{I^s}$ and the noised target image $\mathbf{I^t}_t$ together. However, it's challenging to capture intricate and non-linear transformations from the concatenation operation. 
Instead, we propose to use cross-attention \cite{vaswani2017attention} to better capture information from the source view image $\mathbf{I^s}$ with the help of light field encodings. 
Specifically, we concatenate the source view image $\mathbf{I^s}$ and its light field $\mathbf{L^s}$ together and apply a half U-Net $\mathcal{G}(\cdot)$ to extract features $\{\mathbf{f}_i^s \} = \mathcal{G}([\mathbf{I^s}, \mathbf{L^s}])$. 
We obtain the query $\mathbf{Q} = W_q \mathbf{f}_i^t$ from the target features with learnable weights $W_q$, and the key $\mathbf{K} = W_k \mathbf{f}_i^s$ and value $\mathbf{V} = W_v \mathbf{f}_i^s$ with learnable weights $W_k$, $W_q$. The cross-attention is calculated by: 

\begin{equation}
    \operatorname{Attention}(\mathbf{Q},\mathbf{K},\mathbf{V})=\operatorname{softmax}\left(\frac{\mathbf{Q} \mathbf{K}^T}{\sqrt{d}}\right) \cdot \mathbf{V}. 
\end{equation}
With the cross-attention operation, the target noise image can fully interact with the source image, making it easy to capture non-linear transformation. Moreover, the light field encodings generate similar representations for pixels corresponding to rays that are close to each other.
Thus, the attention operation will prioritize features from corresponding positions. 
In practice, due to the quadratic computation complexity of the attention, we only apply the cross-attention on low-resolution U-Net features. 

% Up Sampler
In image space, we 2x downsample all images to a uniform size, aiming to improve the training efficiency. We upsample the image back to the original resolution during the inference stage. However, a simple bilinear upsampling operation may fall short of preserving the fidelity of the synthesized images. Thus, we introduce a super-resolution module to upsample the synthesized image to the original size. We train a separate DDPM \cite{ho2020denoising} as a refiner to enhance the image quality by addressing potential artifacts and imperfections that may arise during the upsampling process.

\subsubsection{Latent Light Field Diffusion.}
One limitation of image LFD is the difficulty and cost associated with scaling it up to high-resolution images, primarily due to the sequential evaluation process. Hence, inspired by Zero-1-to-3 \cite{liu2023zero1to3}, we introduce latent LFD, which applies LFD in the low-resolution latent space through the fine-tuning of the latent diffusion model (LDM) \cite{rombach2022high}. 
In detail, following \cite{liu2023zero1to3}, we finetune from the Stable Diffusion Image Variation (SDIV) \cite{sdiv}. 
We take the pre-trained auto-encoder $\mathcal{E}(\cdot)$ to compress both source view image $\mathbf{I^s}$ and target view image $\mathbf{I^t}$ into a low-resolution latent code $\mathbf{z^s} = \mathcal{E}(\mathbf{I^s}) \in \mathbb{R}^{h' \times w' \times c}$ and $\mathbf{z^t} = \mathcal{E}(\mathbf{I^t}) \in \mathbb{R}^{h' \times w' \times c}$. As for the light field encodings, we adjust the downsample factor $s$ to make them have the same dimension as the latent code. We denote these low-resolution light field encodings as $\mathbf{'L^s}$ and $\mathbf{'L^t}$. 
Similar to image LFD, we concatenate noise latent $\mathbf{z^t}_t$ and $\mathbf{'L^t}$ together.
Meanwhile, we extract features from source images via a half pre-trained U-Net $'\mathbf{f^s} = \mathcal{G}'( [\mathbf{z^s}, \mathbf{'L^s}])$.  
Besides, SDIV additionally utilizes CLIP \cite{radford2021learning} $\mathcal{H}(\cdot)$ to extract global semantic feature vector from the source image  $ g = \mathcal{H}(\mathbf{I^s})$. To this end, when performing cross-attention, the keys and values come from the concatenation $['\mathbf{f^s}, g]$. 
The training process aims to reconstruct the clean latent code of target image $\mathbf{z^t}_0$. Besides these, the other operations are the same as image LFD. 
Lastly, once we obtain the predicted clean latent code $\hat{ \mathbf{z^t}_0}$, we reconstruct the RGB image with the pre-trained decoder $\mathcal{D}(\cdot)$.

\section{Experiments}
\label{sec:Experiments}

\subsection{Experimental settings}
\subsubsection{Datasets.}

Following the setting in Zero-1-to-3 \cite{liu2023zero1to3}, we train our latent LFD on Objaverse \cite{Deitke_2023_CVPR}, which consists of  800k+ 3D models created by artists. For each 3D model, images and corresponding camera matrices are rendered from 12 different views. We follow the train-test split of Objaverse in \cite{liu2023zero1to3} and resize images to $256\times 256$. At testing time, we randomly pick 700 objects, take the first image as the source view, and generate images in the remaining 11 views accordingly. 
Additionally, we also evaluate the latent LFD in a zero-shot setting. In detail, we evaluate its performance on RTMV \cite{tremblay2022rtmv}: we choose 20 complex scenes containing randomly positioned objects and test our model from $16$ different views.

For image space LFD, we benchmark it on ShapeNet Car \cite{chang2015shapenet}.  
We follow 3DiM \cite{watson2022novel} and resize all images to $128\times 128$. For each scene, we take the 64-th view as the source image and generate the remaining 250 views.

\subsubsection{Evaluation Metrics. }

We evaluate the generation quality between the ground truth image and the synthesized results using four different metrics: Peak signal-to-noise ratio (PSNR), Structure Similarity Index Measure (SSIM) \cite{wang2004image}, Learned Perceptual Image Patch Similarity (LPIPS) \cite{zhang2018perceptual}, and Frechet Inception Distance (FID) \cite{heusel2017gans}. 
% Specifically, PSNR, SSIM and LPIPS mainly evaluate the reconstruction error
SSIM and LPIPS are calculated pair to pair with ground truth images and report the average. PSNR and FID scores are computed with the same number of ground truth views and generated views.

\begin{figure*}[t]
  \centering
   \includegraphics[width=\linewidth]{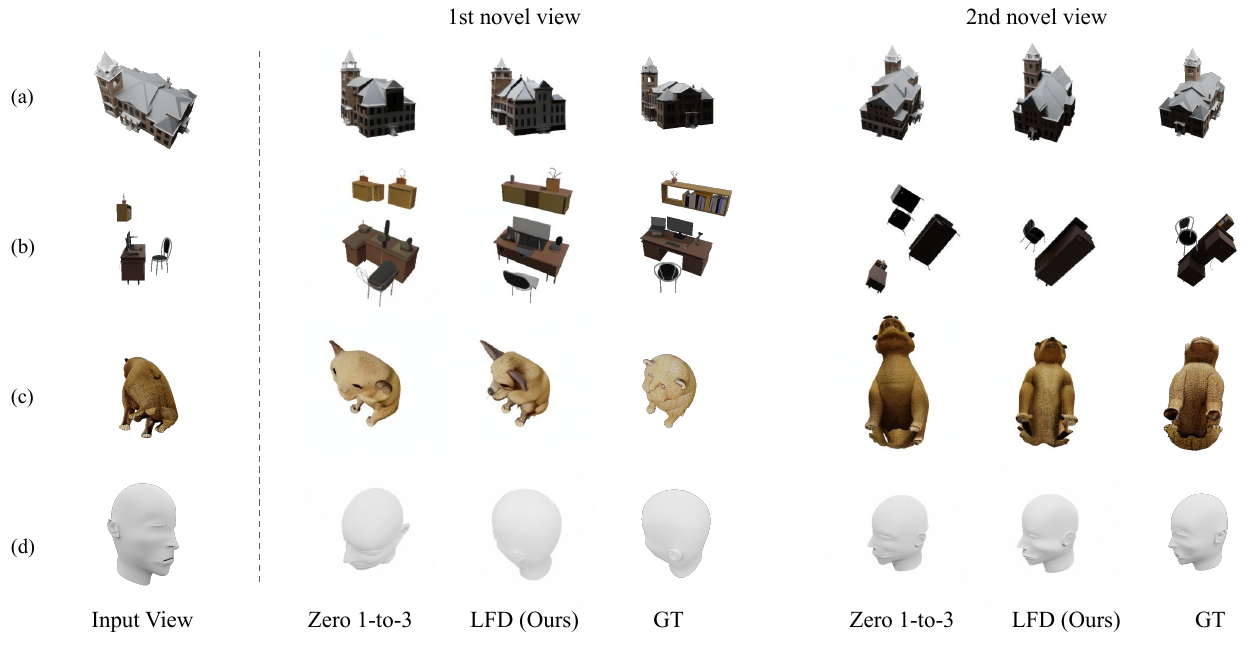}
   \caption{\small{ Comparison of latent LFD and Zero-1-to-3 \cite{liu2023zero1to3} on the Objaverse dataset. For each object, the first image is the input view. We randomly synthesize two novel views. More results and video visualization can be found in Supplementary. }}
   \label{fig:objaverse_compare}
\end{figure*}

\begin{table}[ht!]
  \caption{Results on the Objaverse datasets \cite{Deitke_2023_CVPR}.}
  \centering
  \resizebox{0.5\linewidth}{!}{
  \begin{tabular}{@{}lccccc@{}}
    \toprule
    Method & $\uparrow$ PSNR & $\uparrow$ SSIM & $\downarrow$ LPIPS & $\downarrow$ FID \\
    \midrule
    Zero-1-to-3 \cite{liu2023zero1to3} & 16.788 & 0.874 & 0.117 & \textbf{10.643}\\
    Latent LFD (Ours) & \textbf{17.853} & \textbf{0.881} & \textbf{0.108} & 11.031\\
    \bottomrule
  \end{tabular}
  }
  \label{tab:objaverse_results}
\end{table}

\subsection{Results of Latent LFD}

We first compare our latent LFD with Zero-1-to-3 \cite{liu2023zero1to3} on Objaverse \cite{Deitke_2023_CVPR}. 
The difference between Zero-1-to-3 and latent LFD is that the former takes camera pose matrices as input, while the latter utilizes light field encoding, a 2D spatial pixel-wise representation of camera pose. 
Since Zero-1-to-3 does not evaluate its performance on Objaverse, we evaluate their officially released model weights and calculate metrics by ourselves.
We randomly pick 700 objects from the test split.
The quantitative results are shown in \cref{tab:objaverse_results}. 
Remarkably, LFD outperformed Zero-1-to-3 across most metrics, especially on PSNR and LPIPS, which are pixel-wise comparisons with ground truth images. 
We also present qualitative comparisons in \cref{fig:objaverse_compare}. 
In detail, in \cref{fig:objaverse_compare}(a)(b), latent LFD maintains better consistency with the input view than Zero-1-to-3.
Meanwhile, \cref{fig:objaverse_compare} (c)(d) show that Zero-1-to-3 sometimes cannot guarantee the correctness of camera view whereas we can. 
Thus, these experimental results suggest that utilizing light field encoding of camera pose can effectively enhance consistency with the reference image and ensure accurate viewpoint coherence.

\begin{figure*}[!t]
  \centering
   \includegraphics[width=\linewidth]{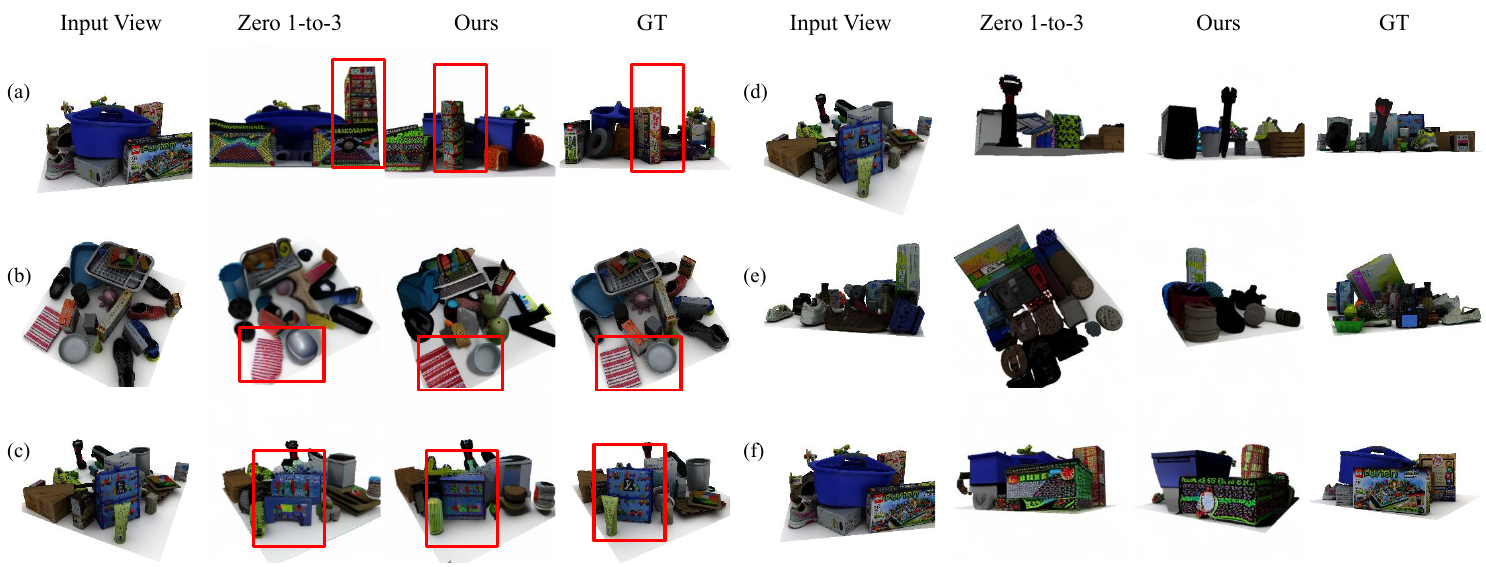}
   \caption{Comparison of latent LFD and Zero-1-to-3 \cite{liu2023zero1to3} on RTMV dataset. }
   \label{fig:rtmv}
\end{figure*}

\begin{figure}[!ht]
  \centering
   \includegraphics[width=\linewidth]{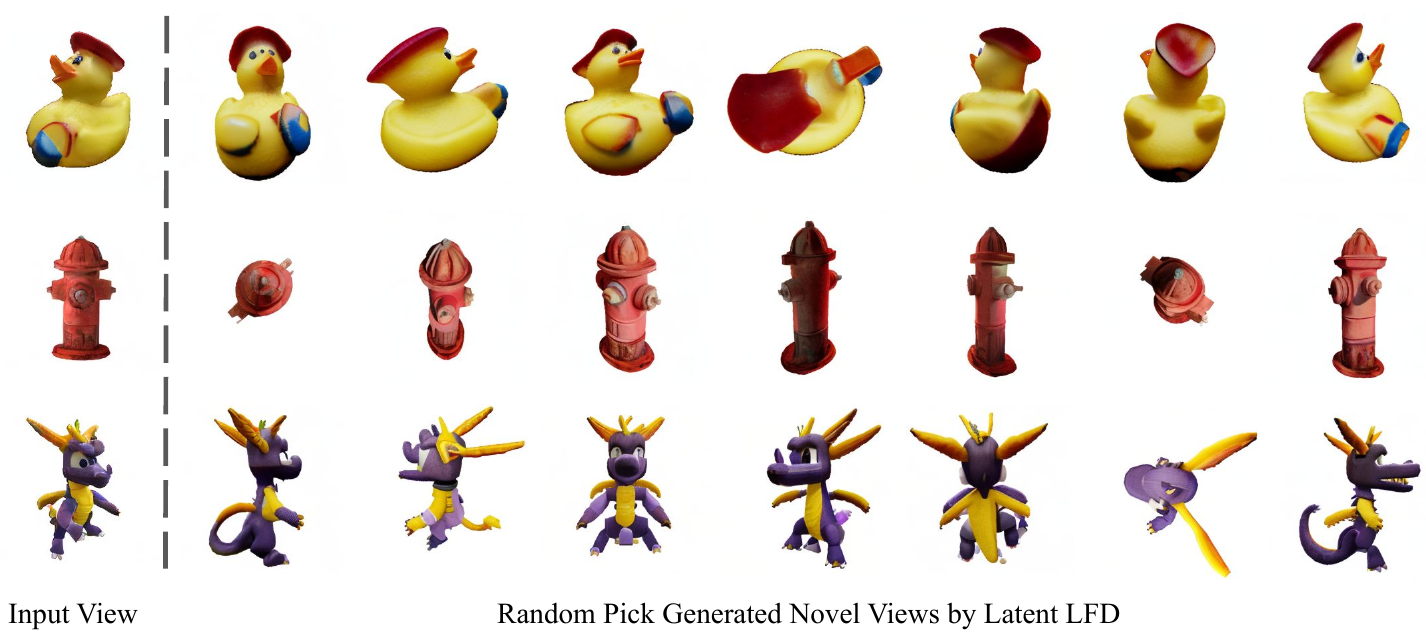}
   \caption{Results of latent LFD on in-the-wild images. The first column displays the input view, while the subsequent columns showcase the diverse and high-resolution novel views generated by latent LFD. Our latent LFD successfully generates images from arbitrary views with consistent appearance. }
   \label{fig:in_the_wild}
\end{figure}

\begin{figure}[!ht]
  \centering
   \includegraphics[width=\linewidth]{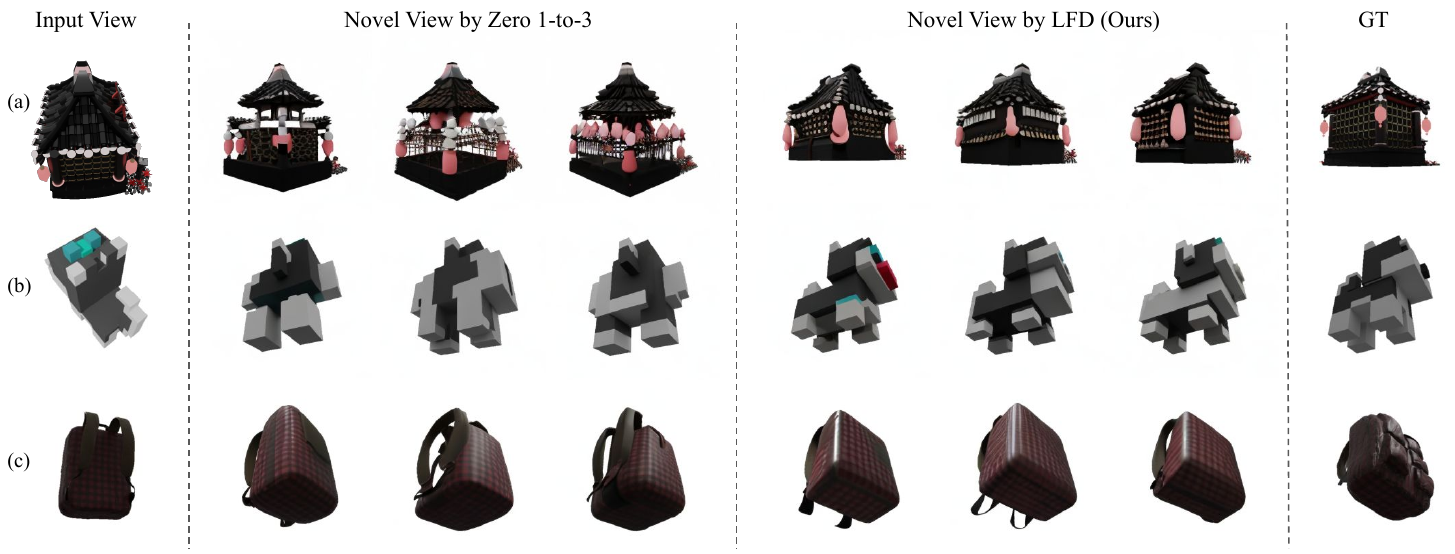}
   \caption{Comparison of diversity. Given an input view and a target viewpoint, we randomly generate 3 samples with different seeds. Zero-1-to-3 \cite{liu2023zero1to3} sometimes cannot guarantee the coherence of viewpoint across different samples. }
   \label{fig:diverse}
\end{figure}

\begin{table}[ht]
  \caption{Quantitative Results on RTMV.}
  \centering
  \resizebox{0.5\linewidth}{!}{
  \begin{tabular}{@{}lccccc@{}}
    \toprule
    Method & $\uparrow$ PSNR & $\uparrow$ SSIM & $\downarrow$ LPIPS & $\downarrow$ FID \\
    \midrule
    DietNeRF \cite{jain2021putting} & 7.130 & 0.406 & 0.507 & 5.143\\
    Zero-1-to-3 \cite{liu2023zero1to3} & 10.405 & 0.606 & 0.323 & \textbf{0.319}\\
    Latent LFD (Ours) & \textbf{12.061} & \textbf{0.625} & \textbf{0.316} & 0.365\\
    \bottomrule
  \end{tabular}
  }
  \label{tab:rtmv_results}
\end{table}

\subsubsection{Zero-shot generalization.}
Following Zero-1-to-3 \cite{liu2023zero1to3}, we further evaluate the zero-shot generalization ability of LFD on RTMV \cite{tremblay2022rtmv} and compare it with both DietNeRF \cite{jain2021putting} and Zero-1-to-3\cite{liu2023zero1to3}. Quantitative comparisons are shown in \cref{tab:rtmv_results}. 
Similarly, latent LFD also achieves state-of-the-art performance across several metrics. 
We also present qualitative comparisons in \cref{fig:rtmv}. 
Specifically,  \cref{fig:rtmv}(a)(b)(c) show our model's ability to accurately ensure better view consistency than Zero-1-to-3 across various aspects, including location (the box in \cref{fig:rtmv}(a)), shape (the bowl in \cref{fig:rtmv}(b)), and integrity (the blue package in \cref{fig:rtmv}(c)). Meanwhile, \cref{fig:rtmv} (d)(e)(f) show our model's ability to ensure better viewpoint coherence. 
Additionally, we also present the performance on in-the-wild images in \cref{fig:teaser} and \cref{fig:in_the_wild}, which illustrates the effectiveness of latent LFD in synthesizing detailed and varied perspectives from out-of-distribution inputs. 
All of these results suggest that our latent LFD maintains better view consistency and viewpoint coherence on out-of-distribution datasets.

\subsubsection{Diversity across samples. }
NVS from a single image is an ill-posed problem. Thus, diffusion models can generate multiple reasonable results. 
We compare the examples' diversity of Zero-1-to-3 \cite{liu2023zero1to3} and latent LFD by generating multiple samples with various seeds.  
Remarkably, Zero-1-to-3 cannot guarantee the coherence of viewpoint across different samples. 
For instance, In \cref{fig:diverse}(c), Zero-1-to-3 put the strap of the backpack on various locations across different samples. 
In contrast, latent LFD produces results that are consistent in geometry while offering diverse appearance information for the invisible parts. 
This also supports that light field encoding can provide better view constraints than camera pose matrices.

\begin{figure}[!t]
  \centering
   \includegraphics[width=\linewidth]{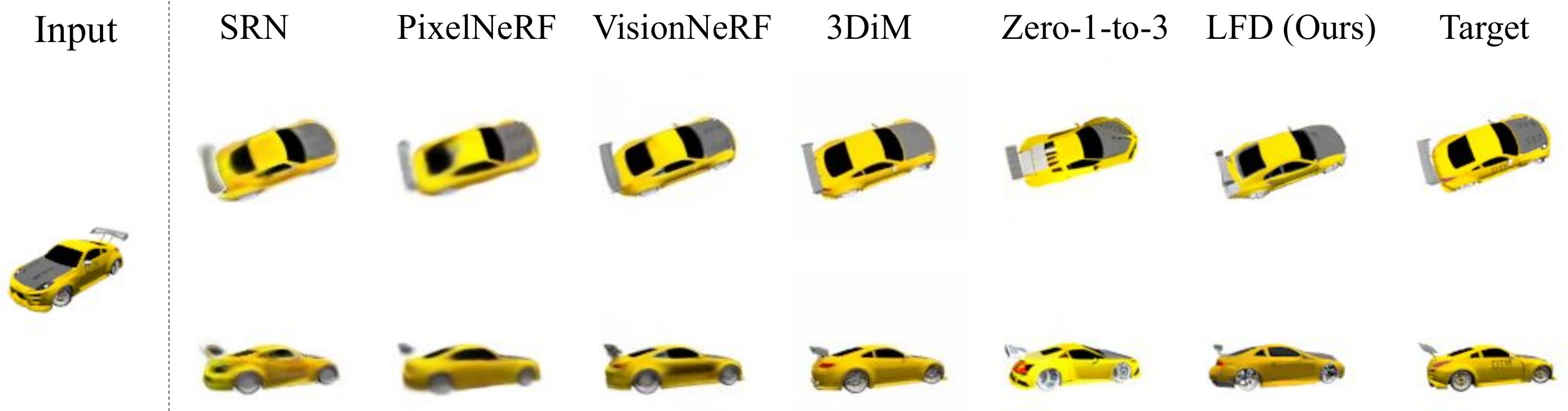}
   \caption{Comparison of image LFD and previous works on ShapeNet Car. }
   \label{fig:car_compare}
\end{figure}

\begin{table}[h]
  \caption{\textbf{Results on ShapeNet Car}. SRN \cite{sitzmann2019scene}, PixelNeRF \cite{yu2021pixelnerf}, VisionNeRF \cite{lin2023vision}, and 3DiM \cite{watson2022novel} baselines are provided by authors of 3DiM \cite{watson2022novel}. }
  \centering
  \setlength{\tabcolsep}{3pt}
  \resizebox{0.6\linewidth}{!}{
  \begin{tabular}{@{}lcccc@{}}
    \toprule
    Method & $\uparrow$ PSNR & $\uparrow$ SSIM & $\downarrow$ FID & Model Size \\
    \midrule
    SRN \cite{sitzmann2019scene} & 22.25 & 0.88 & 41.21 & 70 M\\
    PixelNeRF \cite{yu2021pixelnerf} & 23.17 & 0.89 & 59.24 & 28 M \\
    VisionNeRF \cite{lin2023vision} & 22.88 & 0.90 & 21.31 &  125 M\\
    % \midrule
    
    3DiM \cite{watson2022novel} & 21.01 & 0.57 & 8.99 & 471 M\\
    \midrule
    DDPM+RT & 19.77 & 0.84 & 13.58 \\
    Image LFD (Ours) & 20.17 & 0.85 & 12.84 &  165 M \\
    \bottomrule
  \end{tabular}
}
  \label{tab:car_results}
\end{table}

\subsection{Results of Image LFD}

% \subsubsection{Results on ShapeNet Car.}
We benchmark image LFD on the ShapeNet Car. 
We compare it against regression-based methods including Scene Representation Networks \cite{sitzmann2019scene}, PixelNeRF \cite{yu2021pixelnerf} and VisionNeRF \cite{lin2023vision}, as well as 3DiM \cite{watson2022novel}, a diffusion-based approach conditioned on rotation and translation matrices. 
We also evaluate a baseline model where the light field input in image LFD is replaced with camera pose matrices, while the rest of the architecture remains unchanged, denoting as ``DDPM + RT" for simplicity. 

The quantitative comparisons are summarized in \cref{tab:car_results}. 
First, image LFD outperforms the baseline ``DDPM+RT" across all three metrics, which demonstrates the advantage of light field representation over camera pose matrices in diffusion-based end-to-end NVS networks. 
Second, similar to the findings in \cite{watson2022novel}, NeRF-based methods \cite{yu2021pixelnerf,lin2023vision} achieve better reconstruction errors (PSNR and SSIM) compared to ours, yet they exhibit notably poorer fidelity metrics. 
The reason is that generative methods can easily produce sharp results, while regression methods often yield blurry outputs \cite{watson2022novel}. 
Moreover, 3DiM \cite{watson2022novel} suggests that the capacity of the UNet from DDPM \cite{ho2020denoising} is limited, and thus it yields poor results on NVS. To address this, 3DiM \cite{watson2022novel} proposes the complex X-UNet with 471M parameters to enhance network capacity. However, our image LFD follows the same UNet as DDPM with only 165M parameters, while bringing comparable performance of 3DiM. Thus, light field encoding also helps the learning when the network capacity is limited. 
We present qualitative comparisons in \cref{fig:car_compare}. Given large rotation angles, NeRF-based methods tend to generate blurry images, while our method can generate high-quality samples with better view consistency. 
These results demonstrate that LFD is also effective in image space.

\section{Limitations}
In our study, the latent LFD model demonstrates significant potential for zero-shot novel view synthesis. Yet, it encounters challenges when generating novel views from highly complex, in-the-wild images, such as landscapes. This limitation parallels that of the Zero-1-to-3\cite{liu2023zero1to3} and arises because our training predominantly utilized synthetic data, with a primary focus on novel view synthesis for individual or groups of objects.

While the integration of light field encoding effectively imposes local pixel-wise constraints, it is not without its limitations. Notably, the encoding does not inherently provide explicit depth information, nor does it capture details regarding the light source of the scene. These aspects represent critical avenues for future exploration and enhancement to broaden the applicability and accuracy of our model in more complex and dynamic real-world scenarios.

\section{Conclusion}
\label{sec:Discussion}
We propose Light Field Diffusion, an end-to-end conditional diffusion model for single-view novel view synthesis. 
By translating the camera pose matrices into light field encoding presentation, LFD achieves better view consistency and viewpoint correctness. 
Experiments results on both latent space and image space demonstrate the advantages of our LFD.  
In the future, we plan to explore LFD on one-to-multi novel view synthesis tasks. Overall, our work introduces a new possible way for NVS with diffusion models. We foresee a wider adoption of light field encoding in generative models.

\bibliographystyle{splncs04}
\bibliography{main}

\newpage

\setcounter{section}{0}
\renewcommand\thesection{\Alph{section}}

\section{Experiment Details}

For the light field encoding, we first transform the camera intrinsic matrix and extrinsic matrix to light field $(6\times h \times w)$. Then we apply position encoding to highlight the minimal changes in the light field: $(6 \times h \times w) \rightarrow (180 \times h \times w)$.

\subsection{Latent Light Field Diffusion.}

\subsubsection{Architecture Details}
The Stable Diffusion model \cite{rombach2022high} is originally trained for text-to-image generation. To make the model accept the image as input, we follow Zero-1-to-3 \cite{liu2023zero1to3} and finetune the Stable Diffusion Image Variation \cite{sdiv} on the Objaverse dataset \cite{Deitke_2023_CVPR}. Since the Stable Diffusion model doesn't accept camera information matrices as condition, we modify the model architecture as follows:
\begin{itemize}

    \item The input channel for the U-Net denoiser is changed from 4 to 184  since we concatenate the target light field encoding with the noisy input to the U-Net denoiser. 

    \item We introduce an additional U-Net encoder to extract features from the source image and source light field encoding. The encoder shares the same architecture with the U-Net denoiser encoder but without time embedding and cross-attention.

    \item The extracted feature from the additional encoder is concatenated with the CLIP embedding, serving as a large condition for cross-attention in the U-Net denoiser.
\end{itemize}

Other hyperparameters are set to the same as Stable Diffusion Image Variation: base channel is 320; attention is used when the downsampling factor is in $[4,2,1]$.

\subsubsection{Training Details}
We finetune Stable Diffusion Image Variation \cite{sdiv} on 8 V100 GPUs about 130 epochs, with a total batch size of 192. We use AdamW \cite{loshchilov2018decoupled} as the optimizer with an initial learning rate $1e^{-4}$. We initialize the parameters of the additional U-Net encoder the same as the encoder of U-Net denoiser.

\subsubsection{Inference Details}
During the test stage, we use DDIM \cite{song2020denoising} sampler with $T=50$ and classifier-free guidance \cite{ho2021classifier} scale 3. The time for synthesizing a novel view using a single NVIDIA A6000 GPU is about $2.5$ seconds.

\begin{figure*}[t]
  \centering
    \includegraphics[width=\linewidth]{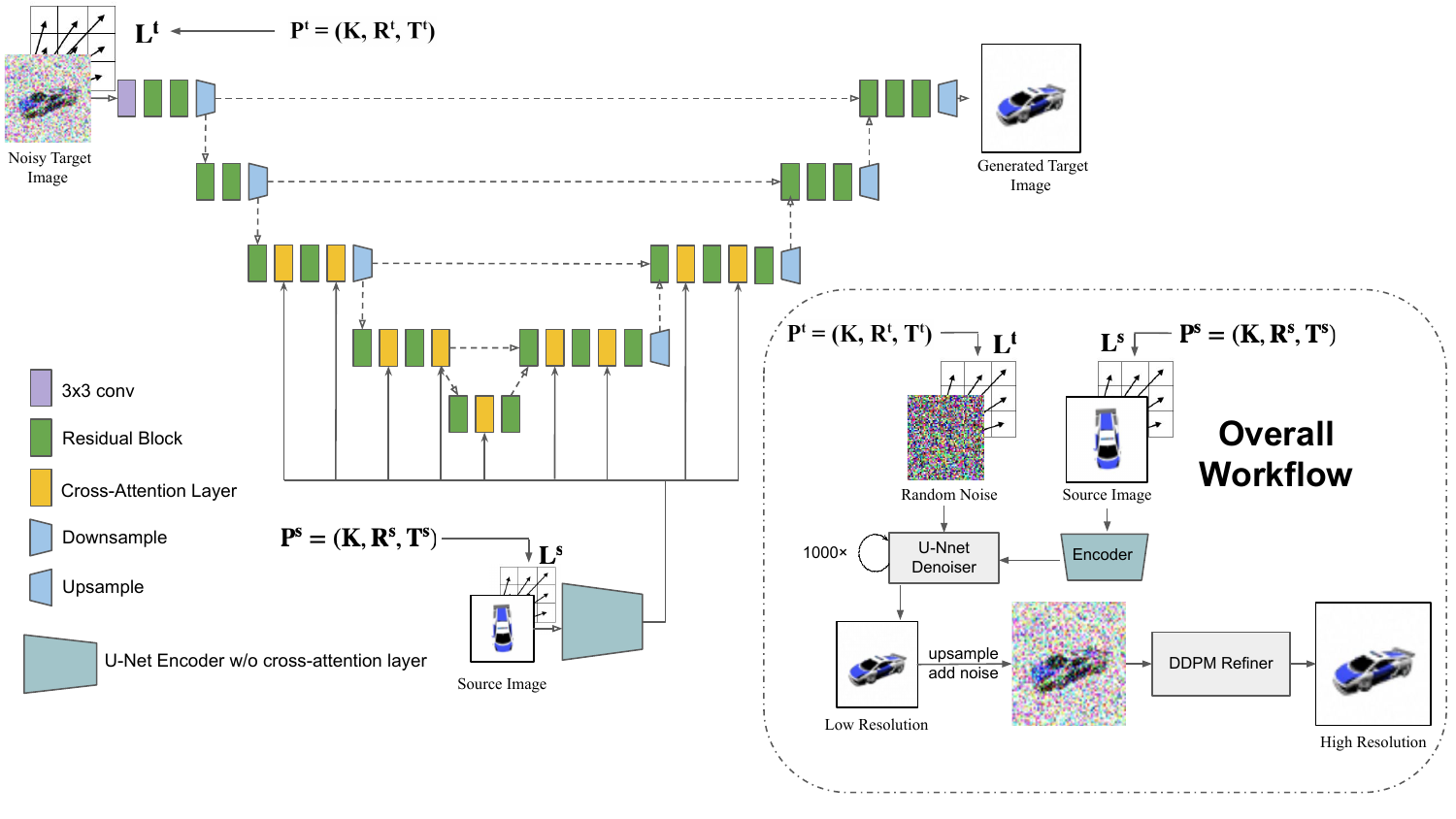}
    \caption{Architecture of image Light Field Diffusion. The LFD transforms the input source camera pose $\mathbf{P^s}$ and target camera pose $\mathbf{P^t}$ into source light field $\mathbf{L^s}$ and target light field $\mathbf{L^t}$. The U-Net takes the concatenation of noisy target image and $\mathbf{L^t}$ as inputs. The separate U-Net Encoder extracts features from the concatenation of the source image and $\mathbf{L^s}$. All images are downsampled to $64 \times 64$ during training. In the inference stage, we use a super-resolution module to upsample and optimize the results from U-Net Denoiser.}
  \label{fig:image_architecture}
\end{figure*}

\subsection{Image Light Field Diffusion.}

\subsubsection{Architecture Details}
\cref{fig:image_architecture} shows the architecture of image Light Field Diffusio and overall workflow.
We build our image Light Field Diffusion based on the standard DDPM \cite{ho2020denoising}. The input channel is also modified to accept the concatenation of target light field encoding and noisy image. A separate encoder that shares the same architecture as the U-Net denoiser encoder but without cross-attention layers is introduced to extract features from the concatenation of source image and source light field encoding.
We apply cross-attention when the downsample factor is 4 and 8.
In image space, we 2x downsample all images to improve the training efficiency. We upsample the generated results back to the original resolution during the inference stage. Instead of a simple bilinear upsampling operation, we introduce a super-resolution module to not only upsample the synthesized images but also preserve the fidelity of generated results.

\subsubsection{Training Details}
We train Light Field Diffusion in image space with $T = 1000$ noising steps and a linear noise schedule $(1e^{-4}$ to $2e^{-2})$ with a batch size of $64$. We utilize AdamW \cite{loshchilov2018decoupled} as the optimizer with an initial learning rate $1e^{-5}$. The training process is performed on a single NVIDIA A6000 GPU.  We train for 550k iterations on ShapeNet Car \cite{chang2015shapenet}. Same as other diffusion-based models \cite{ho2020denoising, rombach2022high}, we apply the exponential moving average (EMA) of the U-Net weights with $0.9999$ decay. We adopt the cross-attention layer when the downsampling factor is 4 and 8. For the super-resolution module, we train a DDPM \cite{ho2020denoising} with $T = 1000$ noising steps and a linear noise schedule $(1e^{-4}$ to $2e^{-2})$ as a refiner.  We utilize Adam as the optimizer with a learning rate $1e^{-5}$ with a batch size of $64$. The training process is performed on a single NVIDIA A6000 for 350k iterations. We also apply the EMA of the U-Net weights with $0.9999$ decay.

\section{Additional Qualitative Results}

\subsection{Latent Light Field Diffusion}
\begin{figure*}[!h]
  \centering
   \includegraphics[width=\linewidth]{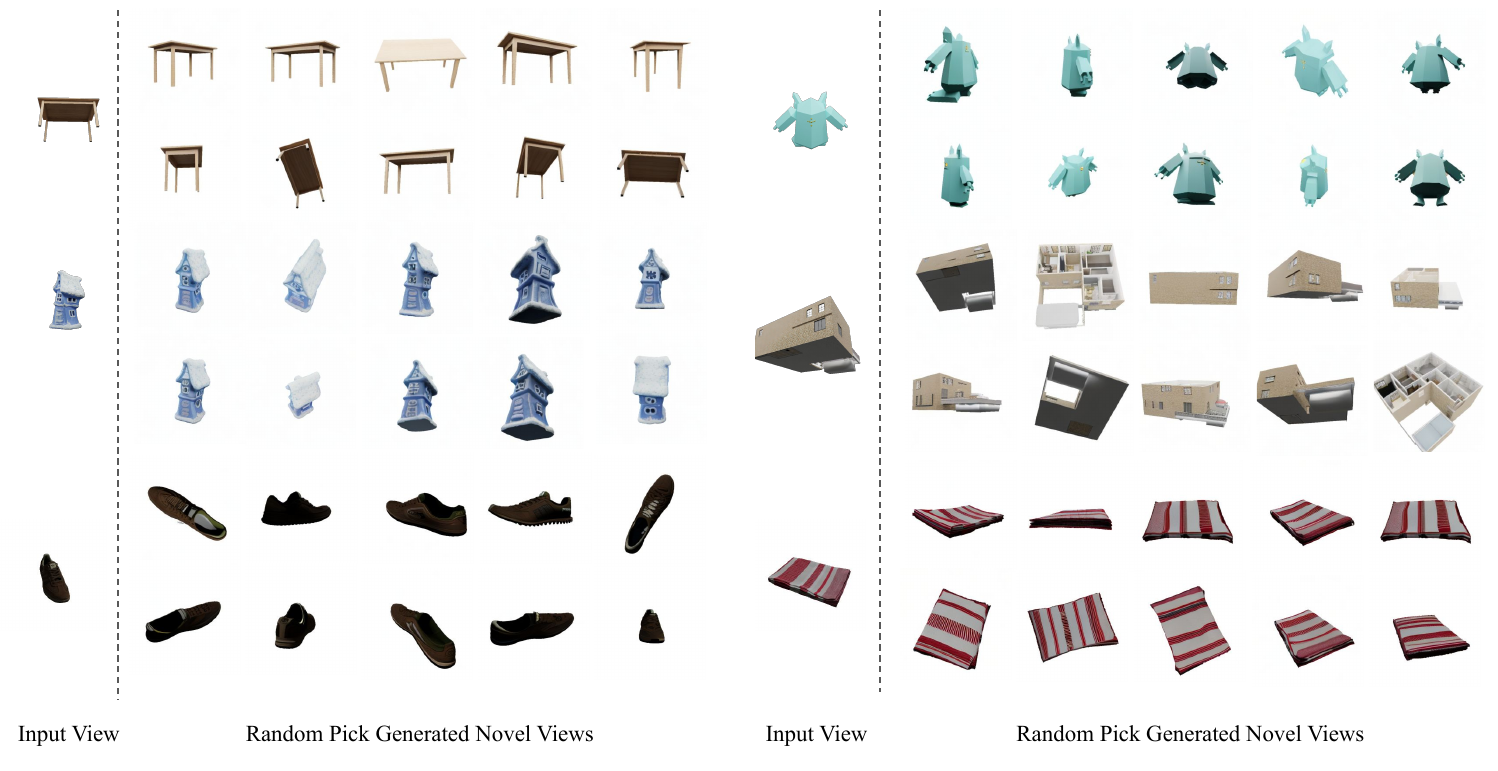}
   \caption{Random sampled novel views on the Objaverse \cite{Deitke_2023_CVPR} and GSO \cite{downs2022google} datasets. The first two rows show results on the Objaverse dataset. The third row shows results on the GSO dataset. We directly show results without cherry-picking.}
   \label{fig:obj_gso_multi}
\end{figure*}

In this section, we present more qualitative results for Latent LFD, showcasing its capability in novel view synthesis across the Objaverse \cite{Deitke_2023_CVPR} and GSO \cite{downs2022google} datasets. \cref{fig:obj_gso_multi} illustrates these novel view synthesis results, highlighting the results' consistency with the reference images. For each object, we randomly generate 10 novel views and directly show the results without cherry-picking.

\begin{figure*}[!h]
  \centering
    \includegraphics[width=1.\linewidth]{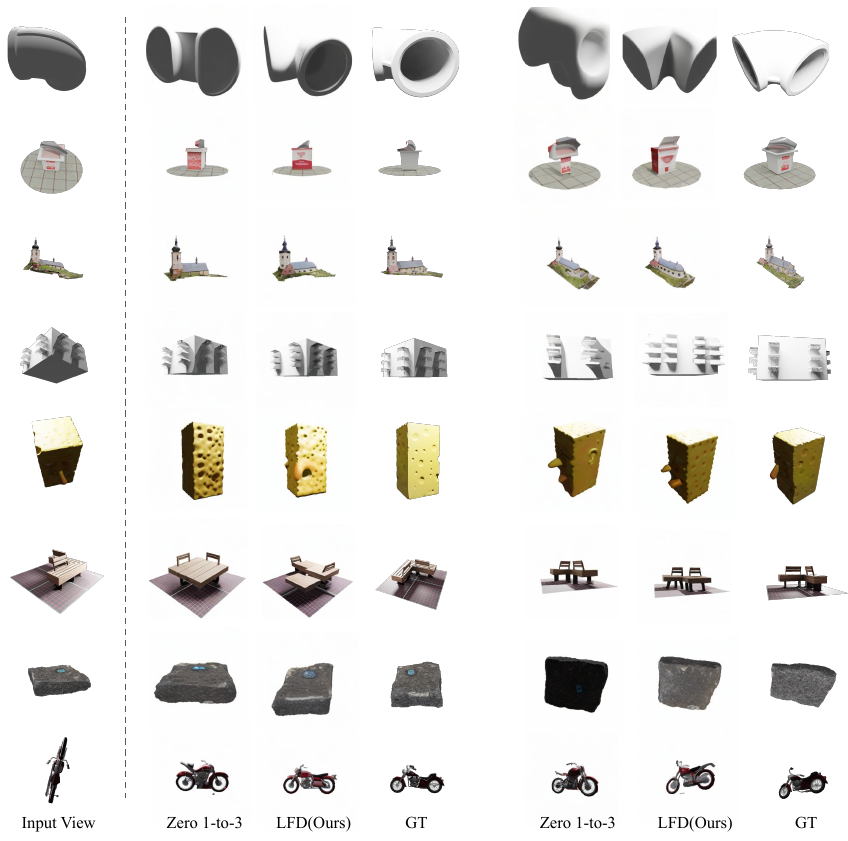}
    \caption{Comparison of Light Field Diffusion and Zero-1-to-3 \cite{liu2023zero1to3} on the Objaverse dataset.}
  \label{fig:obj_compare_1}
\end{figure*}

\begin{figure*}[!h]
  \centering
    \includegraphics[width=1.\linewidth]{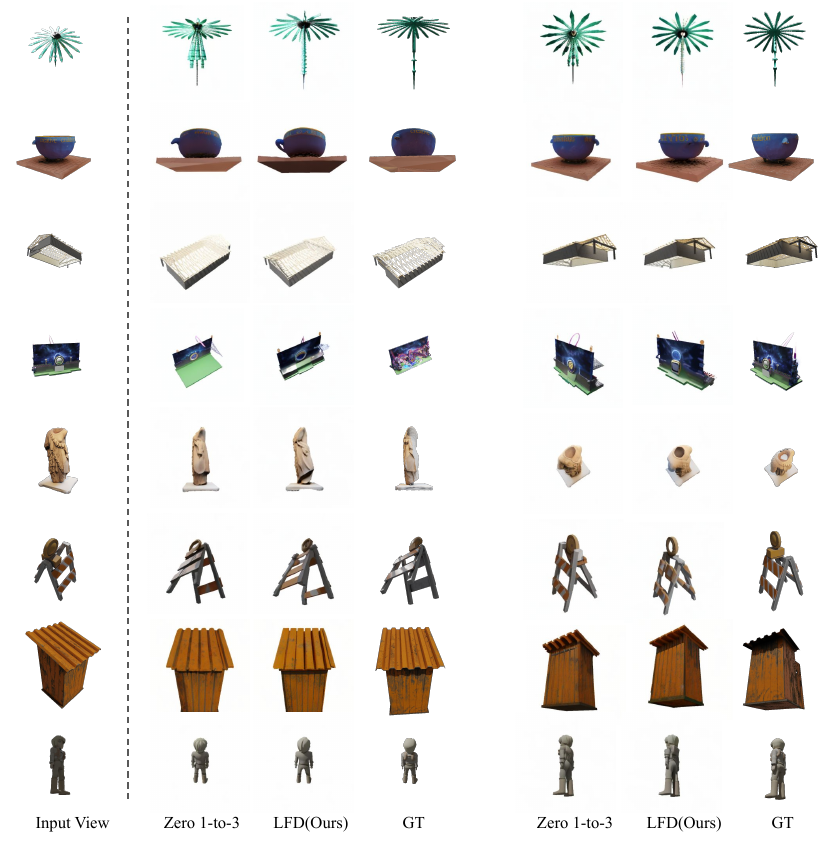}
    \caption{Comparison of Light Field Diffusion and Zero-1-to-3 \cite{liu2023zero1to3} on the Objaverse dataset.}
  \label{fig:obj_compare_2}
\end{figure*}

\begin{figure*}[h]
  \centering
    \includegraphics[width=1.\linewidth]{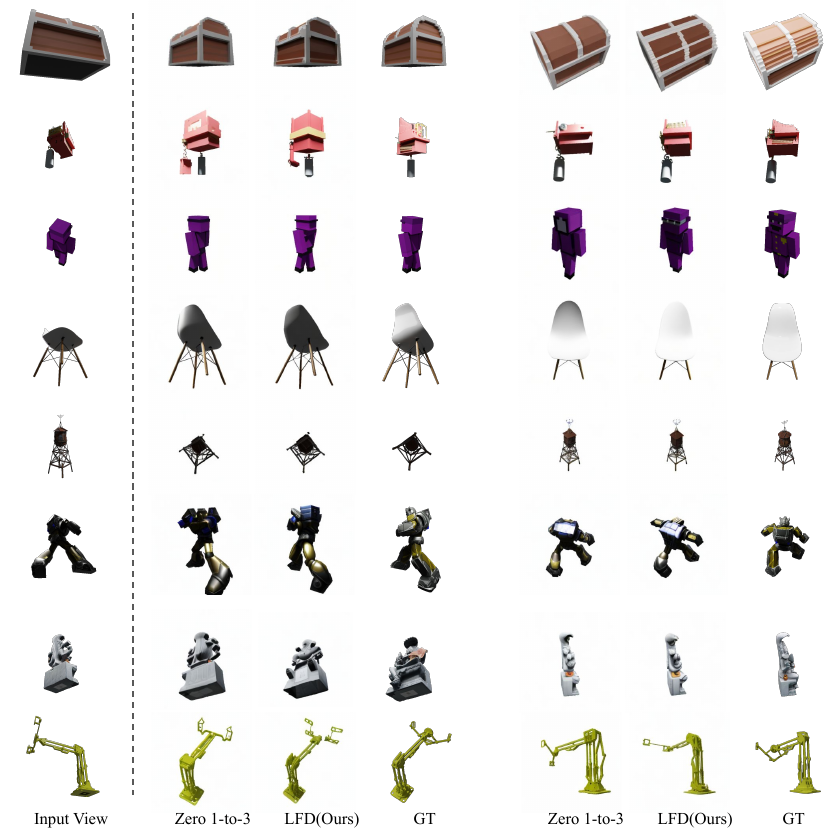}
    \caption{Comparison of Light Field Diffusion and Zero-1-to-3 \cite{liu2023zero1to3} on the Objaverse dataset.}
  \label{fig:obj_compare_3}
\end{figure*}

Next, we extend our comparative analysis of our Latent LFD against the Zero-1-to-3 within the Objaverse dataset. Through figures \cref{fig:obj_compare_1}, \cref{fig:obj_compare_2}, and \cref{fig:obj_compare_3}, we illustrate a side-by-side comparison of novel view synthesis across a diverse selection of objects drawn from the Objaverse test set, underscores our model's versatility across various object categories.

Our comparison demonstrates the superior capability of our latent LFD in producing reasonable results. Notably, our model consistently maintains alignment with the reference images, ensuring accurate viewpoint coherence. The examples provided not only reaffirm the findings discussed in the main paper but also highlight the Latent LFD's advanced proficiency in handling a wide range of object categories, further establishing its potential in novel view synthesis tasks.

\subsection{Image Light Field Diffusion}
\begin{figure}[t]
  \centering
   \includegraphics[width=\linewidth]{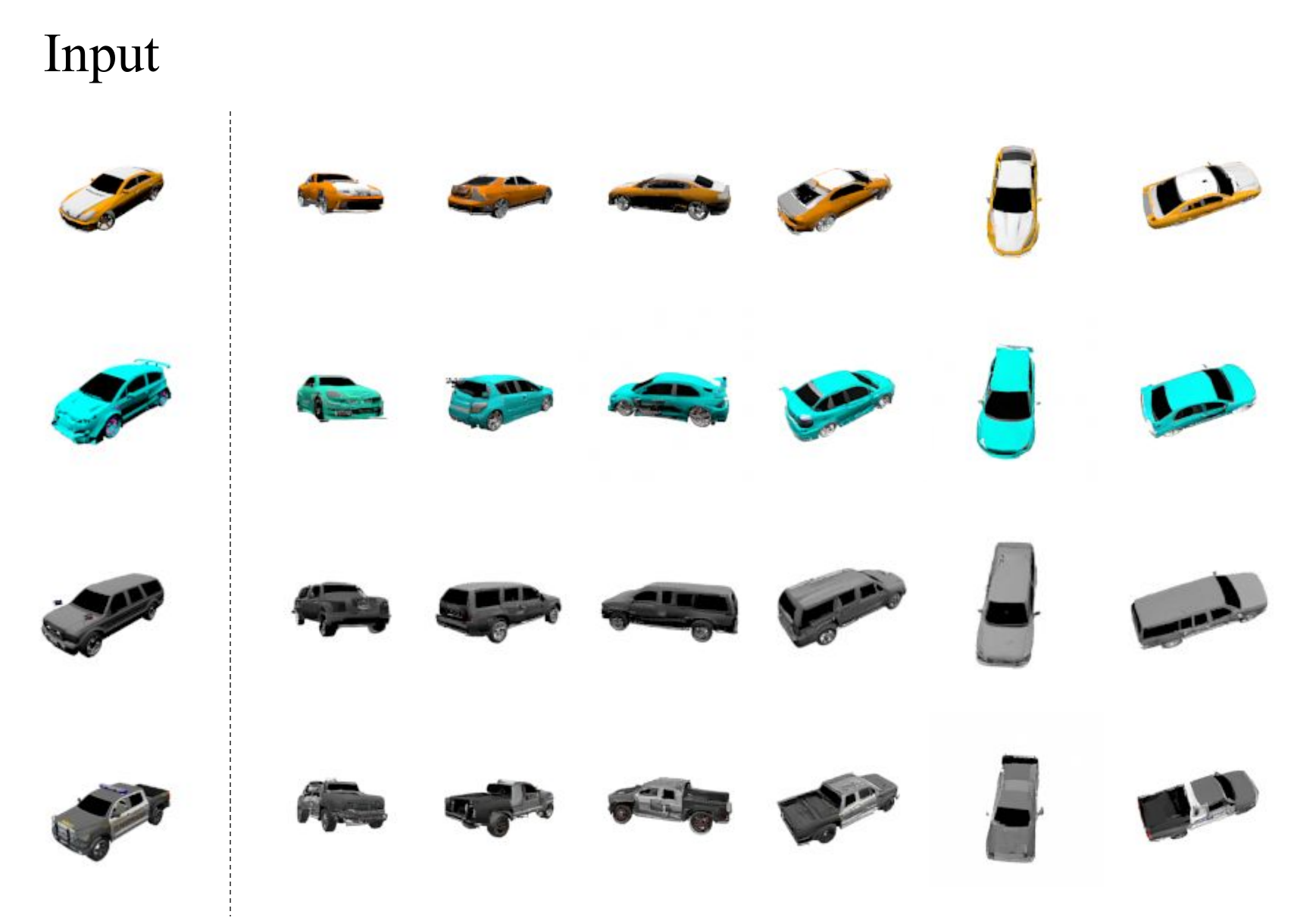}

   \caption{Qualitative ShapeNet Car results. All results are generated based on a single input. We choose 6 different views to present. These results show the consistency of our outputs with reference images.}
   \label{fig:car_all}
\end{figure}

In addition, we present qualitative results for image LFD on ShapeNet Car in \cref{fig:car_all}. This selection of results aims to highlight the model's adeptness at synthesizing images from large view rotation.

\section{Supplement Experiment}
In this section, we provide additional experiments on image Light Field Diffusion, aimed at substantiating the value of the proposed contributions.

\subsubsection{Super-resolution}
We compare the super-resolution module with a naive upsample algorithm in this study to underscore the significance of refining the image subsequent to its generation. Given that our model is trained using a $64 \times 64$ size, employing a basic upsampling technique would result in blurred outcomes. Consequently, the inclusion of a refiner becomes imperative for achieving high image quality. The results are shown in Table \ref{tab:abalation_remove}. With the super-resolution module, our results show improvement in all three metrics.

\subsubsection{Refine Step}
In this study, we conducted trials with the refiner in super-resolution module using different step counts. For a fair comparison, we start with the same $64 \times 64$ low-resolution results. As indicated in Table \ref{tab:refine_step}, considering all three metrics – PSNR, SSIM, and FID – the Light Field Diffusion model utilizing a 200-step refiner emerges as the superior. Therefore, we select 200 as the timestep of the refiner in super-resolution module.

\begin{table}[!ht]
  \caption{\textbf{Abalation Study}. We remove the super-resolution module from Light Field Diffusion and instead use a naive upsample algorithm. We can see a huge deterioration in the FID score. The super-resolution module guarantees high image quality.}
  \centering
  \setlength{\tabcolsep}{4pt}
  \resizebox{0.6\linewidth}{!}{
  \begin{tabular}{@{}lcccccc@{}}
    \toprule
   Method & $\uparrow$ PSNR & $\uparrow$ SSIM & $\downarrow$ FID \\
    \midrule
    LFD w/ super-resolution module & \textbf{20.17} & \textbf{0.85} & \textbf{12.84} \\
    LFD w/ naive upsample algorithm & 20.05 & 0.83 & 51.47 \\
    \bottomrule
  \end{tabular}
}
  \label{tab:abalation_remove}
\end{table}

\begin{table}[!ht]
  \caption{\textbf{Abalation Study}. We test several refiners with different steps. With the increase of timesteps, the FID score decreases. But the PSNR score decreases from 200 steps to 300 steps.}
  \centering
  \setlength{\tabcolsep}{3pt}
  \resizebox{0.6\linewidth}{!}{
  \begin{tabular}{@{}lccccc@{}}
    \toprule
    Method & $\uparrow$ PSNR & $\uparrow$ SSIM & $\downarrow$ FID & Time\\
    \midrule
    LFD w/ 50-steps refiner & 20.09 & 0.83 & 35.35 & 1.01s\\
    LFD w/ 100-steps refiner & 20.14 & 0.84 & 24.84 & 1.95s\\
    LFD w/ 150-steps refiner & 20.17 & 0.84 & 16.93 & 2.91s\\
    LFD w/ 200-steps refiner & \textbf{20.17} & \textbf{0.85} & 12.84 & 3.87s\\
    LFD w/ 300-steps refiner & 20.04 & 0.85 & \textbf{9.55} & 6.33s\\
    \bottomrule
  \end{tabular}
  }
  \label{tab:refine_step}
\end{table}

\end{document}